\documentclass[runningheads]{llncs}
\usepackage[T1]{fontenc}
\usepackage{graphicx}
\usepackage{multirow}
\usepackage{amsmath}
\usepackage{booktabs}
\usepackage[font=small,labelfont=bf]{caption}
\usepackage{url}
\begin{document}

\title{Gait Patterns as Biomarkers: A Video-Based Approach for Classifying Scoliosis}
\titlerunning{A Video-Based Approach for Classifying Scoliosis}

%
%

\authorrunning{Z. Zhou et al.}  
 \author{
   Zirui Zhou\inst{1}$^{\dagger}$ 
   \and Junhao Liang\inst{1}$^{\dagger}$ 
   \and Zizhao Peng\inst{1,2} 
   \and Chao Fan\inst{1} 
   \and \\ Fengwei An\inst{1} 
   \and Shiqi Yu\inst{1}$^{*}$ 
 }
\institute{Southern University of Science and Technology, Shenzhen, China 
\and 
The Hong Kong Polytechnic University, Hong Kong, China
}

%
%
\maketitle              
\newcommand\blfootnote[1]{%
\begingroup
\renewcommand\thefootnote{}\footnote{#1}%
\addtocounter{footnote}{-4}%
\endgroup
}
\begin{abstract}
Scoliosis presents significant diagnostic challenges, particularly in adolescents, where early detection is crucial for effective treatment. Traditional diagnostic and follow-up methods, which rely on physical examinations and radiography, face limitations due to the need for clinical expertise and the risk of radiation exposure, thus restricting their use for widespread early screening. In response, we introduce a novel video-based, non-invasive method for scoliosis classification using gait analysis, effectively circumventing these limitations. This study presents Scoliosis1K, the first large-scale dataset specifically designed for video-based scoliosis classification, encompassing over one thousand adolescents. Leveraging this dataset, we developed ScoNet, an initial model that faced challenges in handling the complexities of real-world data. This led to the development of ScoNet-MT, an enhanced model incorporating multi-task learning, which demonstrates promising diagnostic accuracy for practical applications. Our findings demonstrate that gait can serve as a non-invasive biomarker for scoliosis, revolutionizing screening practices through deep learning and setting a precedent for non-invasive diagnostic methodologies. The dataset and code are publicly available at \url{https://zhouzi180.github.io/Scoliosis1K/}.

\blfootnote{$^{\dagger}$ Equal contribution. \\
$^{*}$ Corresponding author: yusq@sustech.edu.cn.   

}

\end{abstract}
\begin{figure*}[ht]
  \centering
  \includegraphics[width=0.9\textwidth]{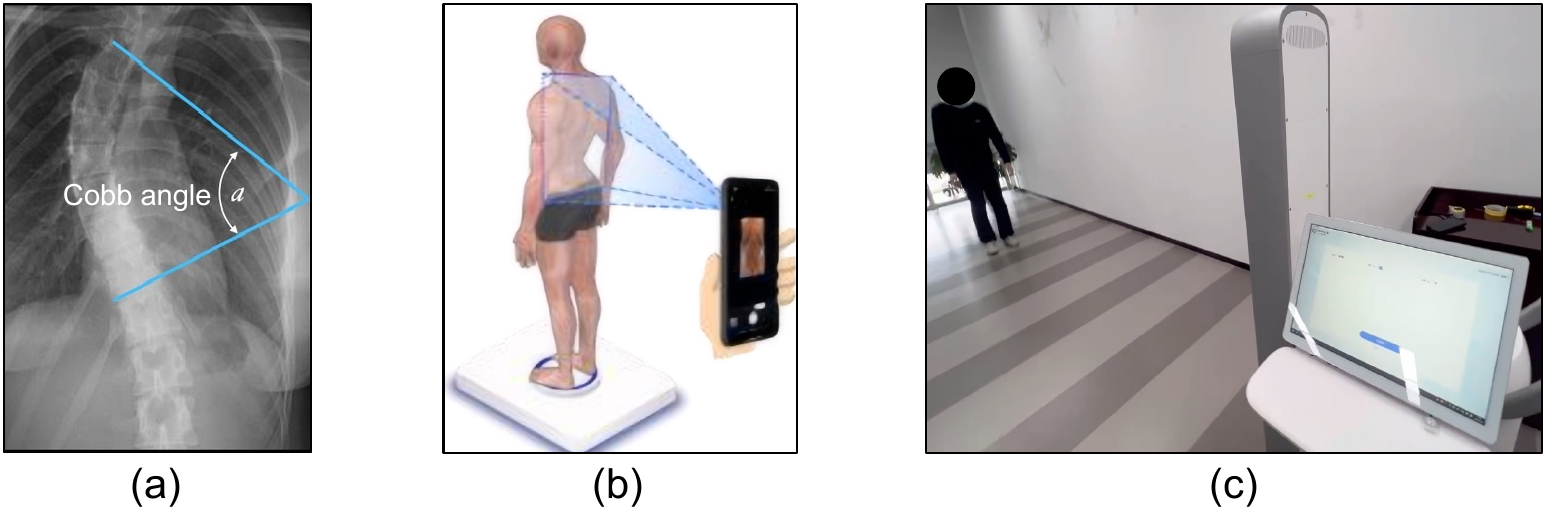}
\caption{Comparative overview of scoliosis diagnosis methods: (a) Traditional X-ray examination, the clinical gold standard~\cite{Thuaimer2024}; (b) Non-invasive analysis of bareback photos~\cite{zhang2023deep}; (c) Our proposed gait analysis approach, enabling efficient, large-scale early adolescent screening to identify cases requiring further radiographic investigation, highlighting its non-invasive and privacy-preserving characteristics.}
  \label{fig:intro}
\end{figure*}
\section{Introduction}\label{sec:introduction}
Scoliosis, a complex spinal deformity characterized by three-dimensional curvature, significantly impacts adolescents' physical well-being and quality of life globally. Clinically, scoliosis is assessed by measuring the Cobb angle, defined as the angle between the upper and lower end vertebrae on standing X-rays (Figure~\ref{fig:intro} (a)), with a threshold exceeding $10^\circ$ indicating scoliosis~\cite{weinstein2008adolescent,reamy2001adolescent}. This early-stage condition is often asymptomatic, potentially leading to severe health issues if undiagnosed~\cite{payne1997does}. In China, scoliosis affects approximately 5.14\% of school-aged children~\cite{hengwei2016prevalence}, highlighting the critical need for effective early screening methods.

Traditionally, scoliosis diagnosis and monitoring have relied on physical exams and radiography, which require significant clinical expertise and expose patients to radiation, limiting early, widespread screening.	Innovations in deep learning have prompted the exploration of non-invasive scoliosis assessment methods, such as bareback photo analysis~\cite{yang2019development,zhang2023deep}. However, these alternatives often raise concerns about privacy and efficiency.	To address these challenges, we introduce a novel video-based method using gait as a biomarker for scoliosis, eliminating the need for direct bodily exposure (Figure~\ref{fig:intro} (b) v.s. (c)). Despite its potential, there is a lack of related work benchmarking this promising direction, primarily due to the absence of public datasets and baseline models.

In response, we developed Scoliosis1K—a groundbreaking dataset featuring over 1k adolescents and 400k frames, setting new standards for video-based adolescent scoliosis screening. To respect privacy and enhance usability, we opted for silhouette data and utilized existing gait recognition techniques~\cite{gaitset,opengait,gaitpart,liang2022gaitedge,fan2023learning} to create ScoNet, demonstrating the viability of gait as a scoliosis biomarker. Furthermore, ScoNet's initial vulnerability to sample imbalance led to the development of ScoNet-MT, an enhanced multi-task learning model.	

This work significantly contributes to the field by (1) creating Scoliosis1K, the first large-scale dataset for video-based scoliosis classification, establishing a new benchmark for the research community; (2) introducing ScoNet, the first baseline model for scoliosis classification through gait analysis, and evolving it into ScoNet-MT to better handle real-world data complexities; and (3) demonstrating that ScoNet-MT exhibits promising diagnostic accuracy for practical applications, underscoring the potential of gait as a non-invasive biomarker for scoliosis and showcasing the transformative impact of deep learning in healthcare diagnostics.

\begin{figure*} [!t]
        \centering
	\includegraphics[width=0.8\textwidth]{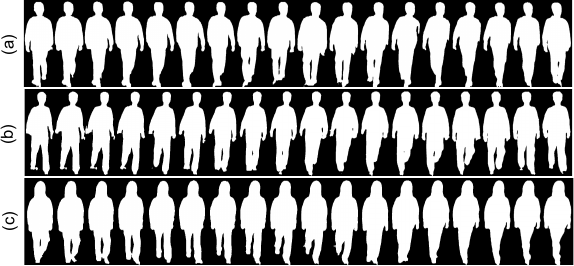}
    \caption{Silhouettes from the Scoliosis1K dataset: (a) positive, (b) neutral, and (c) negative samples.}
	\label{fig:dataset}
\end{figure*}

\section{Dataset}\label{sec:dataset}
\subsection{Overview of Scoliosis1K}
To our knowledge, Scoliosis1K is the first large-scale dataset specifically designed for video-based scoliosis classification. The samples are categorized into three diagnostic groups: positive (Cobb angle $>$ 10$^\circ$), neutral (Cobb angle $\approx$ 10$^\circ$), and negative (Cobb angle $<$ 10$^\circ$). The dataset comprises 1,050 adolescent participants from a middle school in China. It includes 447,900 silhouette images from 1,493 video sequences. Figure~\ref{fig:dataset} shows three sequences representing the different categories. The study received approval from the relevant ethics committee, with informed consent obtained from all participants and their guardians.	

\subsection{Data Collection and Preprocessing}
The videos were captured at 720p resolution using a camera. Participants were instructed to walk along a corridor. The camera was positioned 1.4 to 4.2 meters from the participants. Data collection was designed to simulate a controlled yet natural walking environment conducive to accurate biomechanical analysis.	Each sequence, containing approximately 300 frames at 15 frames per second, was annotated by scoliosis experts. Experts used well-established scoliosis screening methods, including visual assessments and the Adams forward bend test, to classify participants. Experts did not classify participants by watching videos; instead, they assessed the participants' bodies directly. Subsequently, all videos from a participant were labeled according to their assessed category.

Due to the relative scarcity of positive scoliosis cases, diagnosed individuals were encouraged to contribute multiple sequences. This approach helped mitigate potential class imbalance, enhancing the dataset's analytical reliability and robustness.	

During preprocessing, raw video footage was rigorously converted into binary silhouette sequences. Choosing silhouettes serves two primary purposes: anonymizing participant data to safeguard privacy, and focusing the deep learning model on the body region rather than the background. These silhouettes, a fundamental component of our dataset, retain critical gait features while discarding irrelevant background information. A detailed description of silhouette extraction is provided in Section~\ref{sec:preprocessing}.

\begin{table*}[t]
\caption{Statistics of the Scoliosis1K dataset, highlighting diversity in diagnostic categories and participant demographics.}
\centering
\resizebox{\linewidth}{!}{
\begin{tabular}{l|l|lllllll}
\hline
                             & Attributes                 & All             &  & Positive        &  & Neutral         &  & Negative        \\ \hline
\multirow{6}{*}{Scoliosis1K} & Number of Participants     & 1050            &  & 176             &  & 82              &  & 792             \\
                             & Number of Sequences        & 1493            &  & 493             &  & 200             &  & 800             \\
                             & Sex(F/M)                   & 641/409         &  & 113/63          &  & 49/33           &  & 479/313         \\
                             & Age(years, mean $\pm$ std) & 15.2 $\pm$ 1.5  &  & 14.3 $\pm$ 1.0  &  & 14.0 $\pm$ 0.6  &  & 15.5 $\pm$ 1.5  \\
                             & Height(cm, mean $\pm$ std) & 163.2 $\pm$ 8.8 &  & 161.6 $\pm$ 7.1 &  & 161.4 $\pm$ 6.7 &  & 163.7 $\pm$ 9.3 \\
                             & Weight(kg, mean $\pm$ std) & 51.9 $\pm$ 10.7 &  & 48.3 $\pm$ 8.4  &  & 46.7 $\pm$ 7.8  &  & 53.3 $\pm$ 11.1 \\ \hline
\end{tabular}

}

\label{tab:dataset}
\end{table*}
\subsection{Demographics and Dataset Characteristics}

Table~\ref{tab:dataset} provides a demographic and clinical overview of Scoliosis1K, illustrating its comprehensiveness for adolescent scoliosis screening. The dataset's design emphasizes scale and diversity, encompassing a wide array of participant attributes and gait patterns. This variety enhances the dataset's potential for training deep models in scoliosis classification based on gait. 

\subsection{Implications for Scoliosis Research}
Scoliosis1K can advance scoliosis diagnosis research in the following aspects:

\begin{itemize}
    \item \textbf{Scale and Scope:} To our knowledge, it is the first large-scale dataset for automatic scoliosis diagnosis, making computer vision-based scoliosis diagnosis feasible.
    \item \textbf{Innovation in Non-Invasive Screening:} Scoliosis1K provides high-quality, annotated silhouette data, addressing the critical need for non-invasive diagnostic tools in scoliosis screening. This fosters innovation by enabling the exploration of methods that prioritize patient safety and privacy.	
\end{itemize}

    Furthermore, Scoliosis1K bridges a critical gap in the availability of high-quality, annotated images for non-invasive scoliosis screening. This contribution not only catalyzes innovation in healthcare technology but also lays the groundwork for expansive future research in automated scoliosis diagnosis. The dataset paves the way for improving public health, particularly in regions with limited medical services.

\begin{figure*} [ht]
        \centering
	\includegraphics[width=\textwidth]{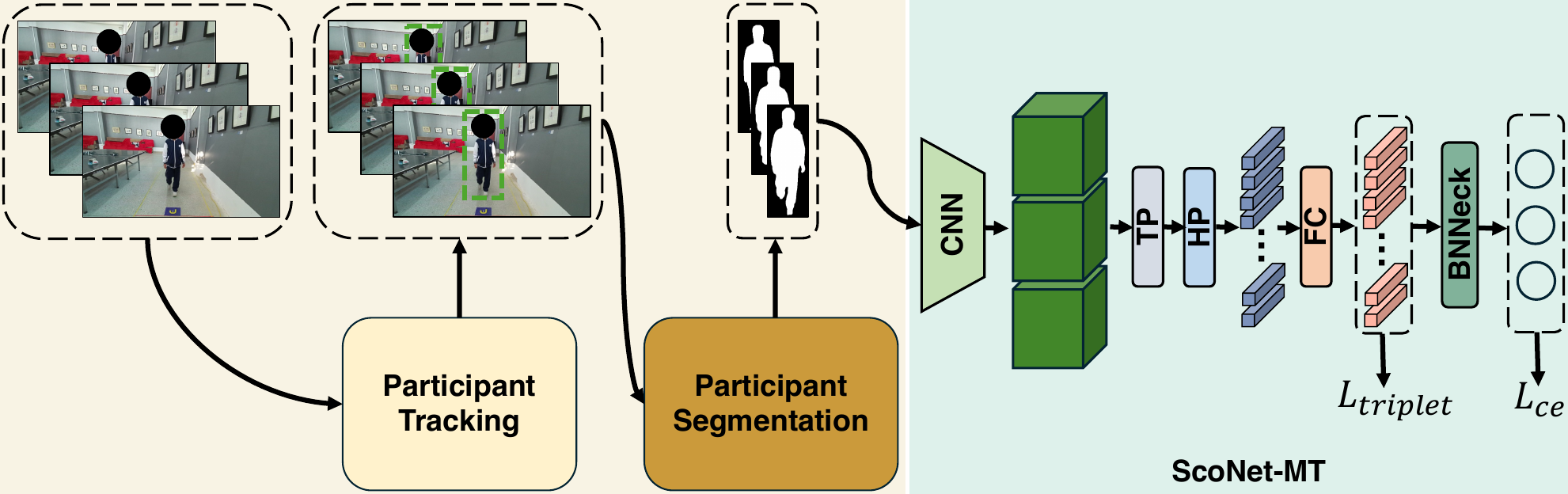}
    \caption{\textbf{The Proposed Pipeline:} The participant is tracked throughout the video, excluding non-participant entities like clinicians. The participant's silhouette is then segmented, followed by scoliosis classification using ScoNet-MT based on gait analysis.}
	\label{fig:pipeline}
\vspace{0mm}
\end{figure*}
\section{Methodology}\label{sec:method}
This section presents our novel approach to scoliosis classification through gait pattern analysis in videos. The method involves three key stages: participant tracking, segmentation, and gait-based scoliosis classification, as shown in Figure~\ref{fig:pipeline}.

\subsection{Participants Tracking and Segmentation}\label{sec:preprocessing}
We used BYTETracker~\cite{bytetrack} for precise person tracking in videos. BYTETracker enhances tracking accuracy by evaluating each detection box, recovering occluded objects, and eliminating irrelevant false detections. It employs tracklet similarity to effectively differentiate between the participant and background, ensuring consistent tracking. 

After tracking, participant segmentation is performed by feeding the cropped image region into PP-HumanSeg~\cite{yuan2020object,liu2021paddleseg}, a robust deep model for precise human segmentation. PP-HumanSeg employs a deep neural network architecture combining an encoder, Spatial Pyramid Pooling Module (SPPM), and Flexible Lightweight Decoder (FLD) to generate binary masks, or silhouettes. We then normalized the silhouettes to a fixed size following gait recognition methods~\cite{iwama2012isir}.

\subsection{Scoliosis Classification based on Gait}
Recognizing gait as a non-invasive biomarker for scoliosis, we introduce ScoNet and its enhanced model, ScoNet-MT, designed to automate the classification process with high efficiency.

\textbf{ScoNet} employs a ResNet-inspired $\mathcal{E}$ architecture to convert participant silhouettes $\mathbf{s}$ into 3D feature maps $\mathbf{f}$:
\[\mathbf{f}=\mathcal{E}(\mathbf{s}) \in R^{n\times c \times h \times w},\]
where $n$, $c$, $h$, and $w$ represent gait frames, channels, height, and width, respectively. Temporal Pooling (TP)~\cite{gaitset} condenses these into essential features $z$: 
\[z=TP({\mathbf{f}}) \in R^{c \times h \times w}.\]
Horizontal Pooling (HP)~\cite{fu2019horizontal} further segments these maps, pooling them into vectors $z_{s}$, with global pooling across 16 horizontal segments for comprehensive feature extraction: 
\[\mathbf{f}^{\prime} = \text{maxpool}(z_{s}) + \text{avgpool}(z_{s}).\]
A fully connected layer then maps these vectors into the metric space, with BNNeck~\cite{bnneck} refining the feature space before final classification using cross-entropy loss:
\[L_{ce} = -\sum_{i=1}^{n} y_i \log(\widehat{y_i}).\]

\textbf{ScoNet-MT} builds on ScoNet by incorporating multi-task learning with a Gait Recognition (GR) task that highlights distinct human motion patterns, reducing bias from non-gait factors. The model uses triplet loss to distinguish subtle gait variations critical for scoliosis classification. In each training batch, $N$ triplets are formed, each comprising an anchor sequence $\mathbf{s}_i^a$, a positive sequence $\mathbf{s}_i^p$ of the same identity, and a negative sequence $\mathbf{s}_i^n$ of a different identity:
\[
L_{triplet} = \sum_{i=1}^{N} \max\left(0, \|f(\mathbf{s}_i^a) - f(\mathbf{s}_i^p)\|_2^2 - \|f(\mathbf{s}_i^a) - f(\mathbf{s}_i^n)\|_2^2 + \alpha\right),
\]
where $f(\cdot)$ is the embedding function transforming each sequence into an embedding space, $|\cdot|_2$ is the euclidean norm, and $\alpha$ is a margin enforcing separation between positive and negative matches. This setup enhances the model’s capacity to discriminate between classes. The total loss function, combining cross-entropy and triplet losses, optimizes ScoNet-MT for accurate scoliosis classification: 
\[
L_{total} = L_{ce} + L_{triplet}.
\]

\section{Experiments}\label{sec:experiments}
\subsection{Setup}
\textbf{Evaluation Protocol.} The dataset was divided into a training set with 745 sequences and a test set with 748 sequences, maintaining a realistic ratio of positive:neutral:negative samples at 1:1:8 in the training set. Specifically, the sequence counts for the three classes are 74, 74, and 596. Model performance was evaluated using three metrics: accuracy, sensitivity, and specificity, defined as follows:
\begin{itemize} \item \textbf{Accuracy:} The proportion of correctly classified samples out of the total number of samples. \item \textbf{Sensitivity:} The proportion of true positives (actual scoliosis cases correctly classified) out of all positive cases. \item \textbf{Specificity:} The proportion of true negatives (actual normal cases correctly classified) out of all negative cases. \end{itemize}

\noindent
\textbf{Implementation Details.} Our models, ScoNet and ScoNet-MT, were implemented using PyTorch~\cite{paszke2019pytorch} and OpenGait~\cite{opengait}, with input image sizes of $64 \times 44$. The training utilized triplet loss with a margin of 0.2. The positive sequence in each triplet was selected from the same gait sequence as the anchor sequence but consisted of different frames. For experiments, 30 frames were selected from each gait sequence as input. All models were trained using an SGD optimizer~\cite{ruder2016overview} with an initial learning rate of 0.1 and a weight decay of 0.0005. The learning rate was reduced by a factor of 10 at 10,000, 14,000, and 18,000 iterations, with training continuing for 20,000 iterations.

\begin{table}[!t]
\begin{minipage}[!t]{.60\textwidth }%
  \caption{Comparison of our method with conventional scoliosis screening techniques. $^*$ indicates results directly cited from~\cite{karachalios1999ten}. The best results are highlighted in bold.}

  \label{tab:method}
  \centering
  
\begin{tabular}{c|ccc}
\hline
Method                                    & Accuracy        & Sensitivity      & Specificity     \\ \hline
Adams Test$^*$~\cite{karachalios1999ten}  & -               & 84.4\%           & \textbf{95.2\%} \\
Scoliometer$^*$~\cite{karachalios1999ten} & -               & 90.6\%           & 79.8\%          \\
ScoNet (Ours)                             & 51.3\%          & \textbf{100.0\%} & 33.2\%          \\
ScoNet-MT (Ours)                          & \textbf{82.0\%} & 99.0\%           & 76.5\%          \\ \hline
\end{tabular}
\hrule height 0pt
\end{minipage}%
~~
\begin{minipage}[!t]{.38\textwidth}
  \centering
  \includegraphics[width=0.8\textwidth]{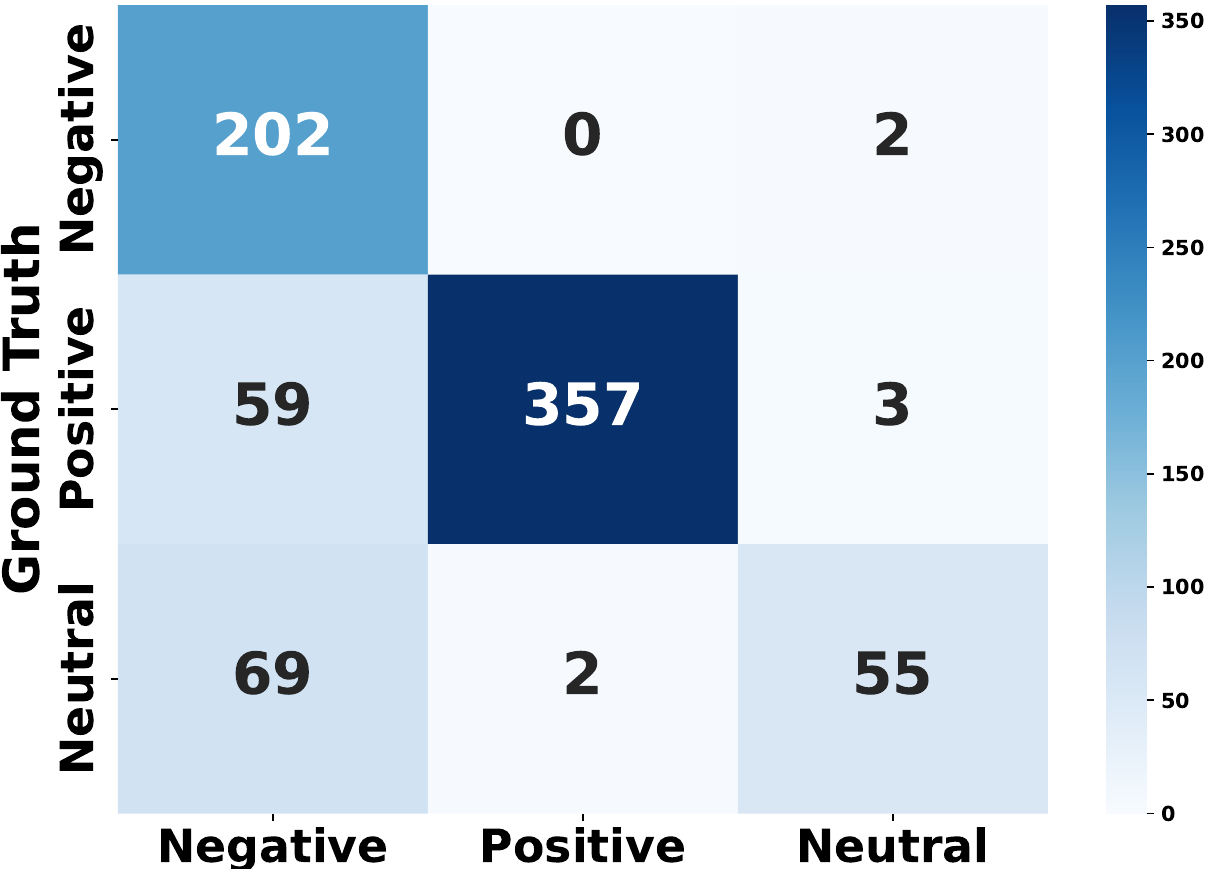}
  \captionsetup{type=figure}
  \caption{Confusion matrix for ScoNet-MT.}
  \label{fig:cm}
\hrule height -12pt
\end{minipage}
\end{table}


\subsection{Results}
ScoNet-MT shows significant improvements in accuracy and specificity compared to the initial ScoNet model, as shown in Table~\ref{tab:method}. The model's accuracy increased by 30.7\%, and specificity by 43.3\%, highlighting the effectiveness of multi-task learning. Although ScoNet-MT's sensitivity surpasses traditional methods like the \textit{Adams forward bend test} and the \textit{Scoliometer}, its specificity, while slightly lower, suggests room for further enhancement. These findings highlight the potential of our approach to surpass human expertise in adolescent scoliosis screening. The confusion matrix (Figure~\ref{fig:cm}) offers detailed insights into the model's precision, particularly in distinguishing between negative cases and others, despite some overlap between positive and neutral classifications.

\begin{figure} [t]	 
  \centering
	\includegraphics[width=0.8\textwidth]{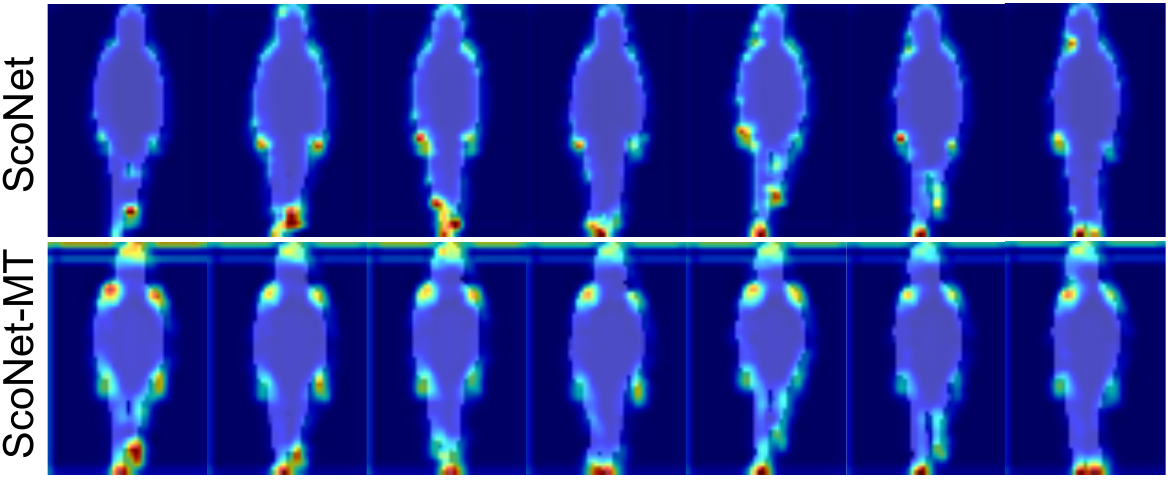}
    \caption{Visualization results of our method.}
	\label{fig:heatmap}
\end{figure}

The heatmaps in Figure~\ref{fig:heatmap} show the regions of interest highlighted by our model using the technique from~\cite{heatmap2017}. While ScoNet mainly focuses on the extremities, ScoNet-MT, through multi-task learning, extends its analysis to critical areas like the head and shoulders, aligning with key motion pattern indicators related to scoliosis identified in the literature~\cite{mahaudens2009gait,kramers2004gait,zhu2021comparison,wen2022trunk}.

\begin{table}[t]
\begin{minipage}[b]{.56\textwidth }%
  \caption{Performance comparison of ScoNet-MT against baseline models and ScoNet. The best results are highlighted in bold.}

  \label{tab:ablation}
  \centering
\resizebox{\linewidth}{!}{\begin{tabular}{l|ccc}
\hline
Method                  & Accuracy & Sensitivity & Specificity \\ \hline
1) Baseline CNN         & 45.5\%   & 99.0\%      & 27.0\%      \\
2) Baseline CNN-Triplet & 56.3\%   & 77.0\%      & 85.0\%      \\
3) Baseline CNN-MT      & 57.8\%   & 96.1\%      & 44.4\%      \\
4) ScoNet               & 51.3\%   & 100.0\%     & 33.2\%      \\
5) ScoNet-Triplet       & 61.2\%   & 100.0\%     & 46.6\%      \\
ScoNet-MT (Ours)        & 82.0\%   & 99.0\%      & 76.5\%      \\ \hline
\end{tabular}}
\hrule height 0pt
\end{minipage}%
~~
\begin{minipage}[b]{.38\textwidth }%
  \caption{Test set accuracy of ScoNet and ScoNet-MT, as influenced by various class distributions (positive, neutral, negative) during training.}

  \label{tab:ratio}
  \centering
\resizebox{0.9\linewidth}{!}{\begin{tabular}{c|cc}
\hline
Pos:Neu:Neg & ScoNet & ScoNet-MT \\ \hline
1:1:2       & 91.4\% & 95.2\%    \\ \hline
1:1:4       & 88.6\% & 90.5\%    \\ \hline
1:1:8       & 51.3\% & 82.0\%    \\ \hline
1:1:16      & 23.7\% & 49.5\%    \\ \hline
\end{tabular}}
\hrule height 0pt
\end{minipage}%
\end{table}

\subsection{Ablation Studies}

\textbf{Baseline Comparisons.} As this is the first video-based solution using a large-scale image dataset and a deep model, we could not find similar methods in the literature for comparison. To demonstrate feasibility and effectiveness, we compared ScoNet-MT with various baseline models in the ablation experiments.
The baseline models include: 
1) \textit{Baseline CNN:} A basic model trained with cross-entropy loss, excluding horizontal pooling and BNNeck. 
2) \textit{Baseline CNN-Triplet:} An improved version of the Baseline CNN, adding triplet loss to enhance category differentiation. 
3) \textit{Baseline CNN-MT:} An extension of the Baseline CNN, incorporating multi-task learning with identity information for better generalization. 
4) \textit{ScoNet:} Our initial model. 
5) \textit{ScoNet-Triplet:} A variation of ScoNet that includes triplet loss to improve feature discrimination.	
Table~\ref{tab:ablation} presents the performance comparison, showing that ScoNet-MT significantly outperforms all baseline models, particularly highlighting the effectiveness of multi-task learning in enhancing diagnostic precision.

\noindent
\textbf{Class-Imbalanced Distribution.} Given the inherent imbalance in scoliosis case distribution, we evaluated the performance of ScoNet and ScoNet-MT across different class ratio settings in the training set. The sequence numbers for these settings were 186:186:373 for approximately 1:1:2, 124:124:497 for 1:1:4, and 41:41:663 for 1:1:16, reflecting real-world scenarios where negative cases predominate.	Table~\ref{tab:ratio} presents our findings, indicating that ScoNet-MT consistently outperforms ScoNet across all ratios. This robust performance, even in highly imbalanced conditions, underscores ScoNet-MT's adaptability and ability to mitigate overfitting, highlighting its superior diagnostic accuracy in realistic settings.

\section{Conclusion}\label{sec:conclusion}
Our study demonstrates the effectiveness of using gait as a non-invasive biomarker for scoliosis. The introduction of the Scoliosis1K dataset, along with the development of the ScoNet and ScoNet-MT models, marks a significant advancement in this field, enabling early and accurate scoliosis classification.	

The implications of our work are far-reaching, offering a scalable and privacy-preserving diagnostic tool with the potential to revolutionize adolescent scoliosis screening, especially in resource-limited regions. Future work will focus on expanding the diversity of our dataset, identifying additional biomarkers, and exploring more effective methods. Developing a mature large-scale scoliosis screening solution using vision-based gait analysis could benefit a vast number of children, particularly in developing countries.

\begin{credits}
\subsubsection{\ackname}
This work was supported in part by the Shenzhen International Research Cooperation Project (Grant No. GJHZ20220913142611021) and  ACCESS (AI Chip Center for Emerging Smart Systems), which is sponsored by InnoHK funding, Hong Kong.

\subsubsection{\discintname}
The authors have no competing interests to declare that are relevant to the content of this article.
\end{credits}

\bibliographystyle{splncs04}
\bibliography{refs}

\end{document}